# CDSA-Net: Collaborative Decoupling of Vascular Structure and Background for High-Fidelity Coronary Digital Subtraction Angiography

Si Li[1], Chen-Kai Hu[2], Zhenhuan Lyu[3], Yuanqing He[1*]

[1]School of Artificial Intelligence and Digital Economy Industry, Guangzhou Institute of Science and Technology, Guangzhou, China.

[2]Department of Cardiology, Second Affiliated Hospital of Nanchang University, Nanchang, China.

[3]School of Artificial Intelligence, Guangzhou Institute of Science and Technology, Guangzhou, China.

**Abstract**

Digital subtraction angiography (DSA) in coronary imaging is fundamentally challenged by physiological motion, forcing reliance on raw angiograms cluttered with anatomical noise. Existing deep learning methods often produced images with two critical clinically unacceptable flaws: persistent boundary artifacts and a loss of native tissue grayscale fidelity that undermined diagnostic confidence. To overcome these limitations, we propose a novel framework termed as CDSA-Net that for the first time explicitly decouples and jointly optimizes vascular structure preservation and realistic background restoration. CDSA-Net introduces two core innovations: (i) A hierarchical geometric prior guidance (HGPG) mechanism, embedded in our coronary structure extraction network (CSENet). It synergistically combines integrated geometric prior (IGP) with gated spatial modulation (GSM) and centerline-aware topology (CAT) loss supervision, ensuring structural continuity. (ii) An adaptive noise module (ANM) within our coronary background restoration network (CBResNet). Unlike standard restoration, ANM uniquely models the stochastic nature of clinical X-ray noise, bridging the domain gap to enable seamless background intensity estimation and the complete elimination of boundary artifacts. The final subtraction is obtained by removing the restored background from the raw angiogram. Quantitatively, it significantly outperformed state-of-the-art methods in vascular intensity correlation and perceptual quality. Most importantly, by delivering artifact-free and physically coherent images, it directly translated to a significant clinical gain. A 25.6% improvement in morphology assessment efficiency and a 42.9% gain in hemodynamic evaluation speed set a new benchmark for utility in interventional cardiology, while maintaining diagnostic results consistent with raw angiograms. The project code is available at https://github.com/DrThink-ai/CDSA-Net.

**Introduction**

Digital subtraction angiography (DSA) removes overlying bone and soft tissue to enable clear arterial visualization. However, its use in coronary interventions is limited by cardiac and respiratory motion, which cause misregistration artifacts [1]. Without a motion-free mask, clinicians must interpret raw angiograms where anatomical noise often masks critical lesions. This results in 22-38% false positive stenosis rates in challenging regions (e.g., low-contrast vessels or rib-overlapped segments) [2], frequently necessitating additional imaging [3] or invasive fractional flow reserve (FFR) [4] that increase time, cost, and radiation.

Early general DSA methods relied on computer vision techniques, such as entropy-based template matching [5], edge-driven registration [6], and invariant feature-based approaches [7]. Targeting coronary motion artifacts, subsequent research developed specialized non-rigid registration B-spline interpolation [8]. Advanced model-based approaches for coronaries included vessel-background separation via independent component analysis [9] and accurate layer extraction using low-rank tensor completion [10]. These methods achieved limited success in the dynamic coronary environment and could not fully eliminate motion artifacts.

Deep learning-based DSA generation has evolved significantly over recent years. Gao et al. [11] used a generative adversarial network (GAN) for single-frame DSA but were limited to static anatomies. Zeng et al. [12] achieved robust coronary subtraction but lost grayscale fidelity. Cantrell et al. [13] leveraged 3D U-Nets with tempor-al inputs for neuro-angiography, not validated for coronary motion. Duan et al. [14] trained on synthetic data that did not fully capture real clinical complexity. Thus, existing methods still suffered from boundary artifacts and loss of grayscale information, that undermined diagnostic confidence.

We developed a novel framework for single-frame coronary digital subtraction angiography (CDSA-Net), which effectively isolated coronary vasculature by removing confounding tissue signals, thereby providing a purified angiographic rendering of coronary structures. The collaborative dual-network explicitly decoupled and jointly optimized vascular preservation and background restoration for high-fidelity subtraction. The final subtraction image was obtained directly by subtracting the raw angiogram from the restored background image.

We employed the transformer-based model as the backbone of our dual-network architecture and introduced two novel key mechanisms to achieve high-fidelity subtraction. One module dedicated to precise vascular structure extraction and the other to seamless background restoration. The main contributions were as follows:

(1) A novel framework for coronary digital subtraction angiography (CDSA-Net): We proposed CDSA-Net, a transformer-based dual-network architecture that achieved coronary angiogram subtraction by restoring background intensity specifically within the vascular regions.

(2) Dual-network enhancement mechanisms: (i) We proposed a novel coronary structure extraction network (CSENet) with a hierarchical geometric prior guidance (HGPG) mechanism. The mechanism combined integrated geometric prior (IGP) with gated spatial modulation (GSM) and centerline-aware topology (CAT) loss supervision, addressing vascular fragmentation and enhancing structural continuity. (ii) We introduced a novel synthetic data generation method and a coronary background restoration network (CBResNet) with an adaptive noise module (ANM) to bridge the domain gap between deterministic restoration and stochastic clinical reality.

(3) Demonstrated superiority in clinical utility: CDSA-Net not only achieved accuracy consistent with the raw coronary angiograms in critical downstream tasks such as coronary stenosis assessment but also enhanced clinical diagnostic confidence and workflow efficiency.

## Methods

### 1. Patient enrollment and data acquisition

This retrospective study included 650 consecutive patients who underwent elective percutaneous coronary intervention (PCI) at the Second Affiliated Hospital of Nanchang University (Nanchang, China) between December 2022 and December 2025. The protocol was approved by the institutional ethics committee, and written informed consent was obtained. Exclusion criteria comprised acute coronary syndromes, chronic total occlusions, and contraindications to contrast. All participants underwent conventional coronary angiography using standardized contrast delivery. The acquisition employed Allura Xper FD20 flat-panel detector system (Philips Healthcare), with 7-10 projection sequences per patient.

### 2. Proposed coronary digital subtraction angiography (CDSA) framework

The dual-network architecture consisted of coronary structure extraction network (CSENet) and a coronary background restoration network (CBResNet). For CSENet, the training set comprised opacified frames from angiograms and their corresponding manually segmented labels. For CBResNet, the training set consisted of pre-contrast frames from the angiographic sequence, and synthetic images generated by superimposing coronary segmentation regions onto these pre-contrast frames. For the whole dataset, 80% of the data was allocated as the training set (with 20% of this subset reserved as the validation set), and the remaining 20% served as the test set.

The proposed CDSA-Net performed automated coronary subtraction in a sequential two-stages process (Figure 1). CSENet first delineated coronary structures, then CBResNet reconstructed background intensity within those regions. Finally, subtracted angiogram was performed by subtracting the raw angiogram from the restored background image. The three pillars included CSENet development, CBResNet

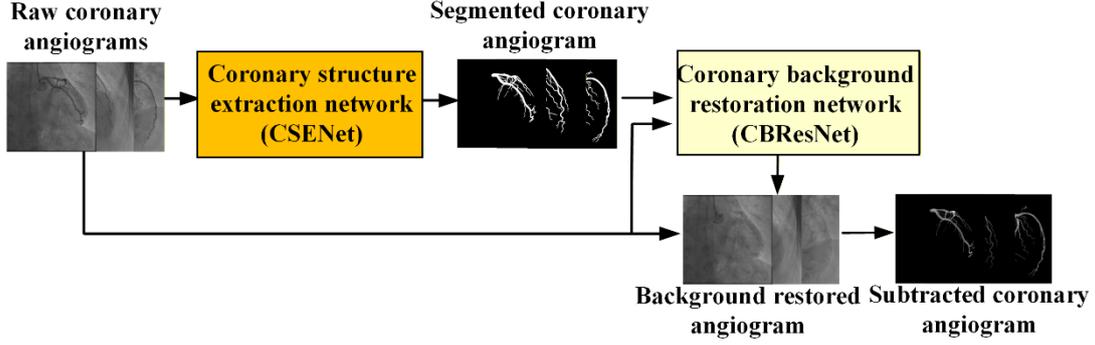

**Figure 1**. Schematic of the proposed coronary digital subtraction angiography (CDSA) method.

development, and comprehensive clinical performance validation, which were described below.

## 3. Coronary structure extraction network (CSENet) development

The development of CSENet model primarily involved three core components (Figure 2(a)): (1) generation of manually annotated coronary structure labels, (2) training of the network and (3) validation of the model performance.

3.1 Generation of manually annotated labels

Two cardiologists annotated coronary boundaries using ITK-SNAP [17] with special attention to clinically critical segments and challenging regions. Labels were quality-controlled via Cohen's $\kappa$-coefficient [18] (greater than 0.85 required) to produce ground-truth masks for supervised learning.

3.2 Training of the proposed CSENet

As shown in Figure 2(b), to overcome vascular fragmentation and low contrast in coronary angiography, we proposed the hierarchical geometric prior guidance (HGPG) framework built upon the MedFormer network [19] as a dedicated geometric perception engine that explicitly modeled coronary topology. Rather than relying solely on data-driven learning, HGPG injected explicit geometric knowledge through three synergistic components: an integrated geometric prior (IGP) extractor, a gated spatial modulation (GSM) mechanism for adaptive feature refinement, and a centerline-aware topology (CAT) loss supervision to enforce structural continuity. This integrated design transformed conventional coronary segmentation into a geometry-constrained reconstruction problem, ensuring that even micro-vessels under severe interference were preserved as connected, physiologically plausible structures essential for clinical quantification.

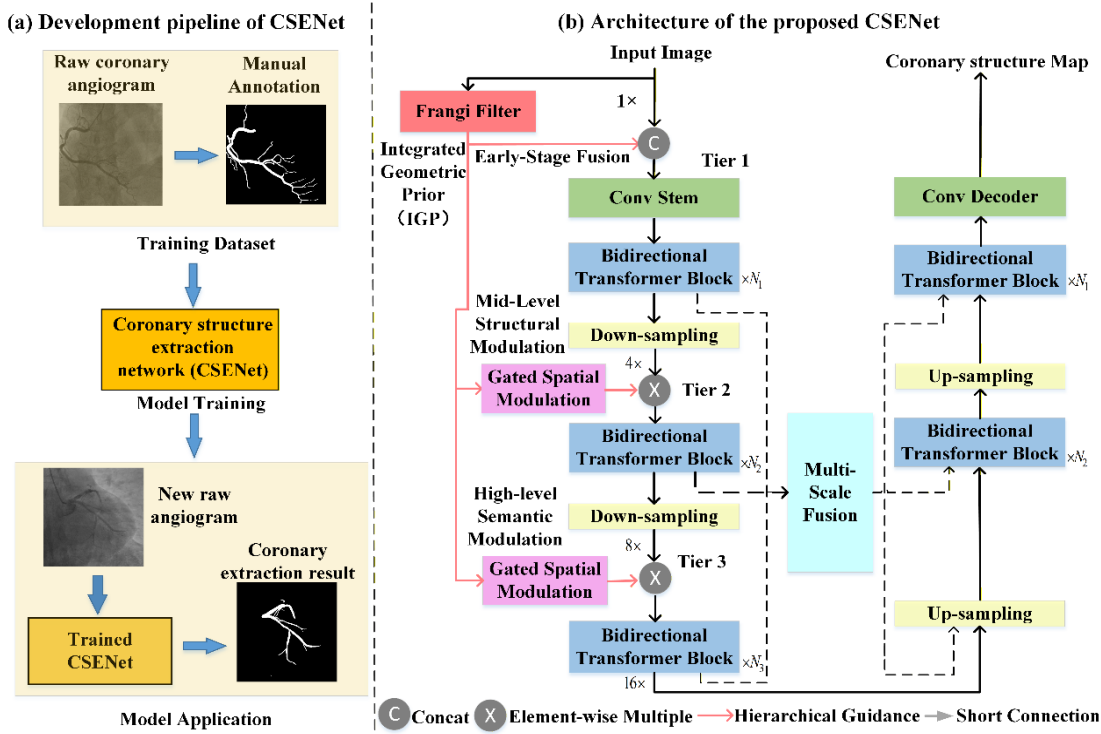

**Figure 2.** Detailed description of the development of coronary segmentation network model. (a) Development pipeline of the model. (b) Architecture of the proposed network.

3.2.1 Integrated Geometric Prior (IGP)

To embed explicit geometric knowledge into the network, we first extracted an Integrated Geometric Prior (IGP) using a multi-scale Hessian-based vesselness filter (Frangi filter) [15], which encoded second-order local intensity structures to capture vascular orientation and scale across diameters ranging from large trunks to fine distal branches. This IGP was then propagated through the MedFormer [19] encoder via a three-tiered hierarchical strategy to maintain structural consistency from pixel-level inputs to deep semantic representations. This hierarchical integration ensured that geometric priors guide feature learning at every stage, transforming a conventional segmentation network into a geometry-aware structure extractor.

3.2.2 Gated Spatial Modulation (GSM) mechanism

Inspired by the success of gated mechanisms in shape-aware segmentation [20], we proposed a Gated Spatial Modulation (GSM) module to adaptively integrate the geometric priors from the scale-space geometric encoder. For the level $l = 2,3$ (shown in Figure 2(b)), a spatial-channel attention gate $\mathcal{G}_1$ was derived:

$$\mathcal{G}_1 = sigmoid\left(BN\left(Conv_{1\times 1}\left(MaxPool(\mathcal{P}_{geom}, kernel)\right)\right)\right) \quad (1)$$

where $sigmoid$ was the nonlinear activation, $BN$ meant the Batch Normalization. $Conv_{1\times 1}$ was the Convolutional operation with $1 \times 1$ kernel. $MaxPool$ meant the Max Pooling with $kernel$ of 2 in this formula 1.

Unlike additive fusion, the GSM module acted as a feature purifier. The features $F_l, (l = 2,3)$ were refined to be $\hat{F}_l$ via a residual gating mechanism, shown in formula 2:

$$\hat{F}_l = F_l \odot (1 + \mathcal{G}_1) \quad (2)$$

where $\odot$ meant the Hadamard Product.

3.2.3 Centerline-Aware Topology (CAT) loss supervision

Standard regional losses like Dice were biased toward thick proximal vessels. We incorporated the clDice (centerline Dice) loss [16] to optimize the topological overlap between the skeletons of the prediction $S_{pred}$ and the ground truth $G_{gt}$. The total objective $\mathcal{L}_{total}$ was:

$$\mathcal{L}_{total} = \lambda_1 \mathcal{L}_{Dice} + \lambda_2 \mathcal{L}_{CE} + \lambda_3 \mathcal{L}_{clDice} \quad (3)$$

where $\mathcal{L}_{Dice}$ was the Dice Loss, $\mathcal{L}_{CE}$ was the cross-entropy loss [19]. $\mathcal{L}_{clDice}$ was the Centerline Dice Loss [16]. $\lambda_1, \lambda_2, \lambda_3$ were parameters of each component.

The network was optimized using stochastic gradient descent [19]. Training leveraged paired inputs of raw angiograms and corresponding expert-annotated segmentation masks. Data augmentation (geometric transformation and signal interference addition) enhanced robustness against clinical variability. Hardware acceleration (NVIDIA A10 GPUs with 24 GB memory) enabled batch size of 8 across 1000 epochs.

3.3 Validation of model performance

Qualitative analysis was performed by three cardiologists using a 5-point Likert scale [21] on the test set (130 patients), assessing vessel continuity, contour accuracy, and specificity in low-contrast, complex, and indistinct regions. Quantitative evaluation employed Dice Similarity Coefficient (DSC) [22], Centerline Dice Coefficient (clDice) [16], 95% Hausdorff distance (HD95) [23] and Intersection over Union (IoU) [24] against expert annotations. Comparisons included UNet++ [25], TransUNet [26], Attention U-Net [27], and the original MedFormer [19].

**4. Coronary background restoration network (CBResNet) development**

The development of the background intensity restoration network model for coronary angiogram primarily comprised three core components (Figure 3(a)): (1) generation of synthetic training image pairs, (2) training of the restoration network, and (3) validation of the model performance.

4.1 Generation of synthetic training image pairs

Here we proposed a dataset synthesis method suitable for the background restoration network. Training data was synthesized by superimposing coronary segmentation mask images (19,734 frames from training set) onto pre-contrast background images (2,251 frames from training set). First, pre-contrast background frames were stratified by coronary territory (left and right coronary angiograms). Synthetic training pairs were generated by superimposing coronary segmentation masks onto pre-contrast backgrounds acquired at the identical projection nomenclature (According to standard clinical protocols, left coronary angiography is performed in six standard projections: AP Caudal, AP Cranial, LAO Cranial, RAO Caudal, RAO Cranial, and Spider View. Right coronary angiography utilizes three projections: AP Cranial, LAO, and RAO. Identical gantry angles are not mandated). This ensured projection consistency and anatomical coherence across all 133,242 synthesized samples.

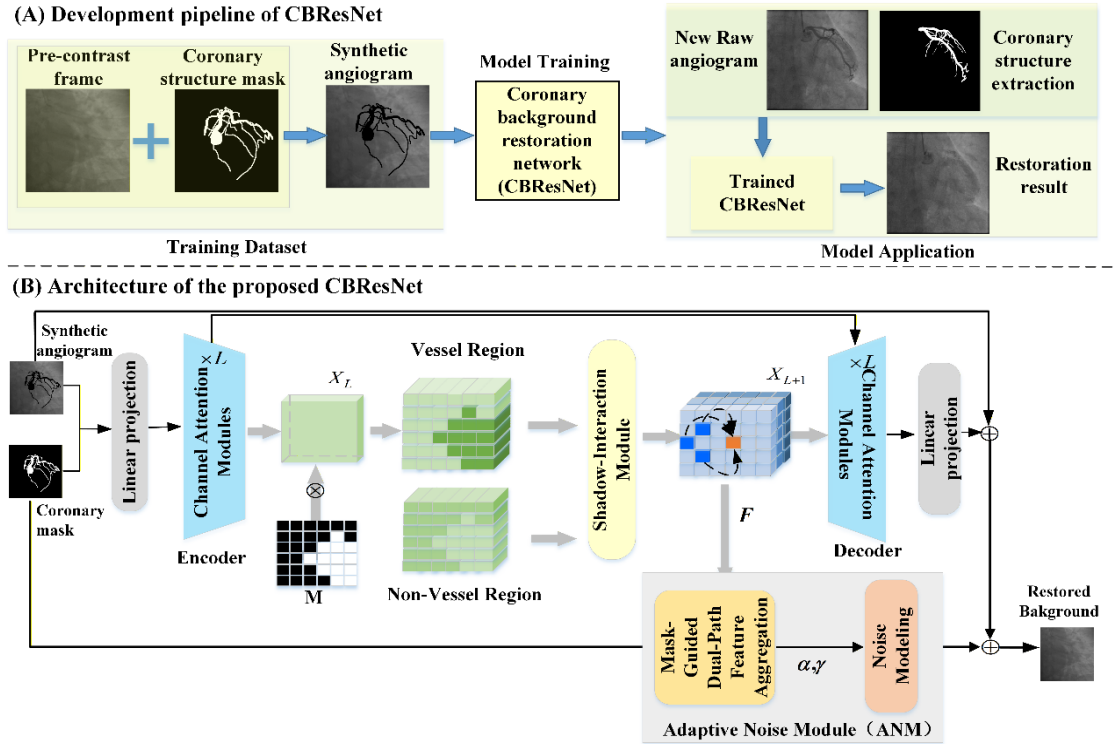

**Figure 3.** Detailed description of the development of background restoration network model. (a) Development pipeline of the model. (b) Architecture of the proposed network.

4.2 Training of the proposed CBResNet

As shown in Figure 3(b), we designed a coronary specific background restoration network (CBResNet) tailored for the unique challenge of recovering native tissue intensities in regions completely obscured by contrast agent, where no direct signal was available. Built upon the ShadowFormer [28] backbone, our key innovation lied in the adaptive noise module (ANM) mechanism. This mechanism explicitly modeled the heteroscedastic nature of clinical X-ray noise (Poisson-distributed quantum mottle and Gaussian-distributed electronic noise). By adaptively adjusting its filtering based on local signal characteristics, ANM enabled seamless integration of restored regions with the surrounding background while preserving natural texture and grayscale continuity essential for clinical interpretation. This noise-aware design transformed a standard restoration pipeline into a physiologically plausible background reconstructor that produced artifact-free, diagnostically reliable subtraction images. The three parts of ANM were described below.

4.2.1 Mask-Guided Dual-Path Feature Aggregation (MG-DPFA)

To isolate noise-relevant features from complex anatomical backgrounds, we introduced a mask-guided dual-path (MG-DPFA) strategy. The dilated vascular mask $M_{dil}$ served as a spatial prior to guide the feature pooling, ensuring that the noise estimation was conditioned on the region of interest (ROI). There were two formulas:

$$f_{mean} = \frac{\sum(CoordAtt(\mathcal{F}) \odot M_{dil})}{\sum M_{dil} + eps} \tag{4}$$

$$f_{max} = max(CoordAtt(\mathcal{F}) \odot M_{dil}) \tag{5}$$

where $\mathcal{F}$ was the high-level latent features from the network bottleneck, $CoordAtt$ meant the coordinate attention [29]. The resulting descriptor $f_{combined} = [f_{mean} \parallel f_{max}]$ was mapped through decoupled linear heads to regress the physical parameters $\alpha$ and $\gamma$ that described below. This aggregation prevented the 'zero-noise collapse' in low-signal-to-noise ratio scenarios, ensuring the model remained sensitive to subtle vessel variations.

4.2.2 Physics prior driven heteroscedastic noise modeling

In the clinical practice of DSA, the raw acquisition was inherently corrupted by signal dependent artifacts. To bridge the domain gap between deterministic image restoratio-n and stochastic clinical reality, we formulated the noise synthesis as a heteroscedastic process. This model explicitly accounted for the dual-source nature of X-ray imaging: quantum mottle (Poisson-distributed shot noise) and electronic stationary noise (Gaussian-distributed read noise) [30].

Given a noise-free reconstructed background $I_{bg}$, the synthesized observation $I_{syn}$ was generated via formula 6:

$$I_{syn} = I_{bg} + \left(\epsilon \cdot \sqrt{\alpha \cdot I_{bg} + \gamma}\right) \odot M_{soft} \tag{6}$$

where $\alpha$ and $\gamma$ presented the predicted gain and stationary noise coefficients, respectively. The term $\epsilon \sim \mathcal{N}(0,1)$ denoted a standard normal distribution. $M_{soft}$ was a spatially-smoothed vascular mask that constrained the noise injection to the anatomically relevant regions.

4.2.3 Statistical moment alignment loss function

To model stochastic X-ray noise, we used a statistical moment alignment loss ($\mathcal{L}_{stat}$) that aligned local mean and variance between the synthesized output $I_{syn}$ and the clinical target $I_{tar}$ in logarithmic domain as a variance stabilizing transformation.

To estimate the localized noise intensity, we employed a sliding window $\Omega$ of size $K \times K$ (typically 11×11). First, as shown in formula 7, we calculated the spatial average of all pixel values within its surrounding neighborhood.

$$\mu_{i,j} = \frac{1}{K \times K} \sum_{m,n \in \Omega_{i,j}} I_{m,n} \tag{7}$$

Second, the entire image was squared element-wise. We computed the local average of these squared intensities as formula 8:

$$E[X^2]_{i,j} = \frac{1}{K \times K} \sum_{m,n \in \Omega_{i,j}} I_{m,n}^2 \qquad (8)$$

Last, the estimated local variance map was obtained by formula 9:

$$\sigma_{i,j}^2 = E[X^2]_{i,j} - (\mu_{i,j})^2 \qquad (9)$$

$\mathcal{L}_{stat}$ was defined as formula 10:

$$\mathcal{L}_{stat} = \sum_{i,j} W_{i,j} \left| \log(\sigma_{syn(i,j)}^2 + \varepsilon) - \log(\sigma_{tar(i,j)}^2 + \varepsilon) \right| + \lambda W_{i,j} |\mu_{syn(i,j)} - \mu_{tar(i,j)}| \qquad (10)$$

Where the spatial weight map $W$ can be set to be $M_{dil}$, $\lambda$ was a hyperparameter, $\varepsilon$ was a small numerical stability constant.

The network was trained with the AdamW optimizer [28] using synthetic data, affine augmentations, and Poisson noise injection for 500 epochs on NVIDIA A10 GPUs (batch size of 8).

4.3 Validation of model performance

Three cardiologists evaluated restored backgrounds against pre-contrast frames using a 5-point Likert scale [21] for overall quality, restoration fidelity, transition integrity, and edge artifacts. Quantitative assessment used peak signal to noise ratio (PSNR) [31], structural similarity index (SSIM) [32], Fréchet inception distance (FID) [33]. Comparisons included ShadowDiffusion model [34] and original ShadowFormer [28].

**5. Overall coronary subtraction**

The Lambert-Beer law [35] established that X-ray attenuation followed an exponential decay. Let $I_0$ denoted the incident X-ray intensity, $\mu_{bg}$ and $d_{bg}$ represented the attenuation coefficient and thickness of background tissues, and $\mu_c$ and $d_c$ denoted the attenuation coefficient and thickness of the contrast agent. The X-ray intensities of the background image $I_{bg}$ (without contrast agent) and the raw image $I_{raw}$ (with contrast agent) were given by:

$$I_{bg} = I_0 exp^{-\mu_{bg}d_{bg}}, I_{raw} = I_0 exp^{-(\mu_{bg}d_{bg}+\mu_c d_c)} \qquad (11)$$

To eliminate the contribution of background structures, subtraction must be performed in the logarithmic domain:

$$I_{DSA} = ln(I_{bg}) - ln(I_{raw}) = \mu_c d_c \qquad (12)$$

In practical digital imaging systems, the pixel intensity (grayscale value) PV was typically linearly proportional to the X-ray intensity (PV $\propto$ $I$). Therefore, logarithmic subtraction could be directly applied to the image grayscale values:

$$PV_{DSA} = ln(PV_{bg}) - ln(PV_{raw}) \qquad (13)$$

The resulting values were linearly proportional to the contrast agent concentration, forming the physical foundation of digital subtraction angiography.

We evaluated our method against the only available deep learning-based coronary subtraction method with functional code (Zeng et al. [11]) using the held-out test set with the complete coronary subtraction pipeline.

5.1 Qualitative comparison

Since no ground truth existed for clinical subtraction images, we conducted a blinded reader study comparing our method with Zeng [11]. Three cardiologists rated each image on a 1-5 Likert scale [21] across four criteria: vessel clarity, vessel integrity, background suppression effectiveness, and boundary artifact severity. Ratings were averaged across readers and cases to obtain final scores.

5.2 Quantitative evaluation

We designed two indirect quantitative evaluations to objectively assess subtraction quality.

5.2.1 Segmentation-based metric

A simple threshold segmentation operation was applied to subtracted images generated by each method. The Dice coefficient [22] between the resulting segmentations and manual annotations on raw angiograms were calculated. Higher Dice scores indicated better structural preservation, as high-quality subtraction would enable more complete and accurate vessel delineation.

5.2.2 Grayscale fidelity assessment

First, pixel-wise Pearson correlation between subtracted and raw angiogram was computed within vessel segments with similar backgrounds. This metric evaluated grayscale consistency, acknowledging that perfect correlation was neither expected nor required due to residual background superposition. Second, Fréchet inception distance (FID) [33] was employed to assess overall perceptual quality, measuring how realistically the subtracted images resembled raw angiograms. Higher correlation indicated better intensity preservation, while lower FID reflected superior perceptual fidelity.

**6. Clinical performance evaluation**

Three experienced interventional cardiologists evaluated raw angiograms and CDSA-subtracted images in randomized order of 130 consecutive test patients.

To evaluate the lesion morphology characteristics, experts annotated lesion location (AHA coronary artery segmentation standard [36]) and stenosis severity. To evaluate the hemodynamic parameters, experts annotated Thrombolysis In Myocardial Infarction frame count (TFC) [37]. All readers specifically evaluated performance in predefined clinically challenging scenarios: low-contrast zones, skeletal overlap regions. Quantitative assessment of above diagnostic concordance involved deviations analysis from CDSA-subtracted to the raw angiograms for 3 continuous parameters (lesion segments number, stenosis severity, TFC value).

To evaluate the efficiency, experts counted the cost time for 2 parameters (stenosis detection time, TFC decision time). Time was quantified as interpretation duration per patient, measured from data loading to final diagnostic confirmation.

To evaluate the clinical value of subjective interpretation in the subtracted angiogram, three cardiologists were invited to perform a blinded evaluation of the raw images, the subtracted images, and the segmented images. The evaluation was conducted a 5-point Likert scale [21] across three domains: diagnostic confidence, image interpretability, and vessel visibility.

**Results**

**1. Coronary structure extraction network (CSENet) development**

Figure 4 showed representative results by several methods. (a) was raw angiogram, (b) was ground truth, (c) to (f) were UNet++, TransUNet, Attention U-Net, MedFormer, (g) was the proposed CSENet. Row 2, 4, and 6 were magnified views of rows 1, 3, and 5, respectively. Magenta regions indicated false positives, while blue regions indicated false negatives in these visualizations. Our method achieved lower false det-ections in high-noise regions (shown in rows 1 and 2). Rows 3 and 4 demonstrated that our method extracted vascular structures more accurately in low-contrast areas. Furthermore, rows 5 and 6 illustrated that the proposed method possessed strong discriminative ability against rib structure.

**Table 1.** Qualitative and quantitative comparative analysis of different segmentation networks.

|  |  | Overall quality | Low contrast | Complex vascular geometries | Indistinct boundary |
|---|---|---|---|---|---|
| Qualitative analysis | UNet++ | 4.253 | 4.417 | 4.579 | 4.182 |
|  | TransUNet | 4.096 | 4.260 | 4.262 | 4.262 |
|  | Attention U-Net | 4.411 | 4.417 | 4.420 | 4.340 |
|  | MedFormer | 4.568 | 4.575 | 4.579 | 4.500 |
|  | Proposed | **4.726** | **4.733** | **4.658** | **4.738** |
|  |  | DSC (%) | clDice (%) | HD95(mm) | IoU (%) |
| Quantitative analysis | UNet++ | 89.89 | 90.13 | 12.2898 | 81.81 |
|  | TransUNet | 89.57 | 89.96 | 10.6855 | 81.24 |
|  | Attention U-Net | 89.65 | 89.85 | 16.2952 | 81.37 |
|  | MedFormer | 90.83 | 90.93 | 7.5069 | 83.37 |
|  | Proposed | **91.84** | **93.18** | **4.2752** | **85.00** |

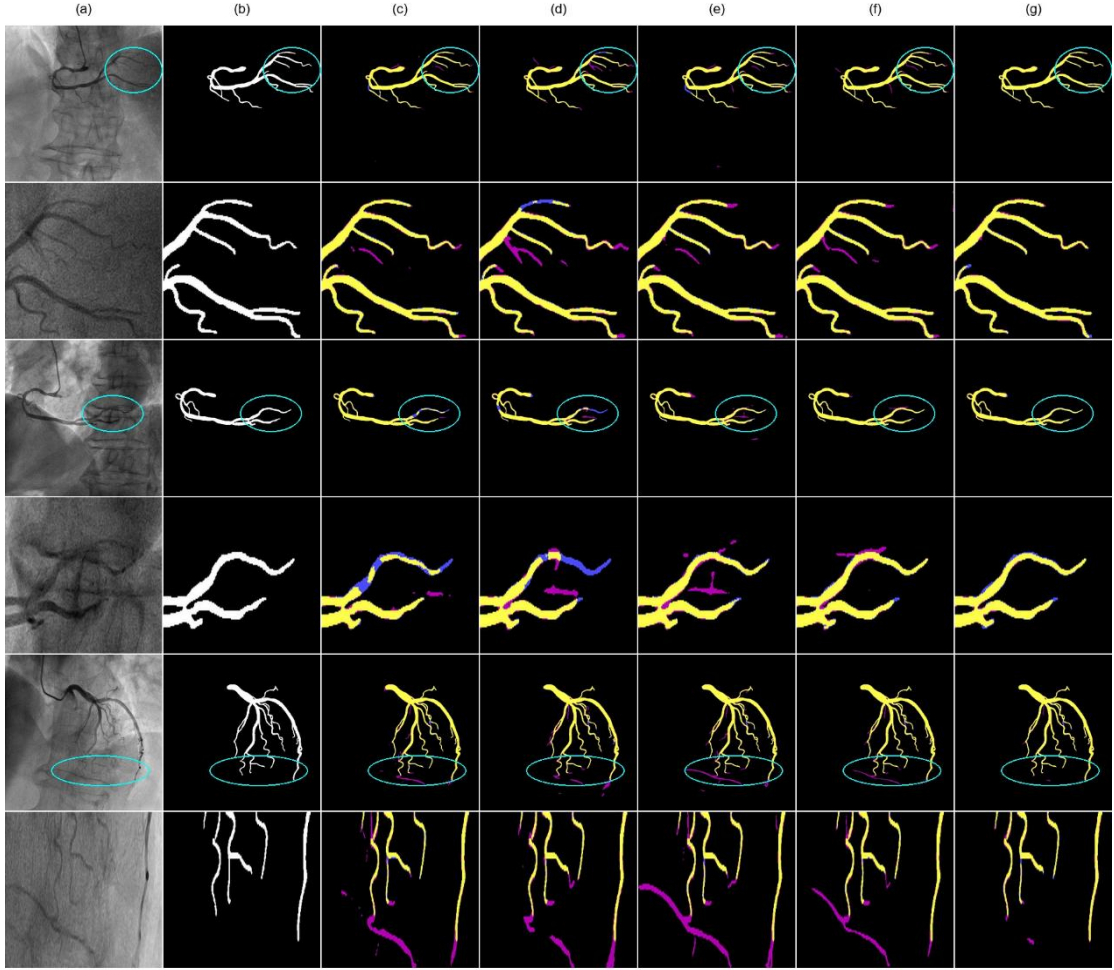

**Figure 4.** Comparison of segmentation results by different networks. (a) raw angiogram; (b) ground truth; (c) UNet++; (d) TransUNet; (e) Attention U-Net; (f) MedFormer; (g) Proposed CSENet.

As shown in Table 1, the qualitative Likert-scale evaluation highlighted the clear advantages of the proposed method across all four perceptual regions. The improvements were especially pronounced in overall quality and boundary sharpness, where it substantially surpassed both the baseline MedFormer and other state of the art networks. This demonstrated that the integrated geometric priors effectively enhanced visual fidelity under challenging imaging conditions.

As shown in Table 1, the proposed method consistently outperformed all competing approaches across all quantitative metrics. It achieved the highest DSC (91.84%) and IoU (85.00%), outperforming the baseline MedFormer by 1.0% and 1.6% respectively, validating the geometric guidance modules. Notably, it attained the highest clDice (93.18%), markedly surpassing MedFormer [19] (90.93%) and other methods (all below 90.2%), demonstrating effective preservation of vascular topology critical for stenosis assessment. Moreover, our method achieved the lowest HD95 (4.28mm), a 40% reduction from MedFormer (7.51 mm), confirming precise boundary delineation even in challenging regions. Collectively, these results underscored the robustness of in achieving accurate region coverage, precise boundary delineation, and preserved vascular topology for clinically applicable coronary segmentation.

## 2. Coronary background restoration network (CBResNet) development

Figure 5 showed representative background restoration results. (a) was synthetic frame, (b) was pre-contrast frame, (c) was ShadowDiffusion, (d) was ShadowFormer, (e) was the proposed CBResNet result. Row 2, 4 and 6 were magnified views of rows 1, 3, and 5, respectively. Compared methods showed distinct boundaries between restored regions and background, whereas our method more closely approximated the physical characteristics of the true background.

As shown in Table 2 (upper), the proposed method was superior across all perceptual metrics, especially in transition integrity and restoration fidelity, indicating seamless blending and preserved details. Unlike ShadowDiffusion [34] and ShadowFormer [28], our results more closely matched the physical characteristics of true coronary anatomy.

As shown in Table 2 (lower), although the proposed method did not improve PSNR or SSIM over ShadowFormer, it still achieved results of relatively high quality. The ANM achieved a VFID of 0.48, representing an 80.9% and an 78.0% improvement over ShadowFormer (2.52) and ShadowDiffusion (2.18). This demonstrated superior vascular texture realism while maintaining background fidelity (NVPSNR 74.30 dB, NVSSIM 0.9999) comparable to ShadowFormer. In contrast, ShadowDiffusion prod-uced a higher global FID of 41.48 and degraded background metrics (NVPSNR 52.48), indicating hallucinations obscured clinical details.

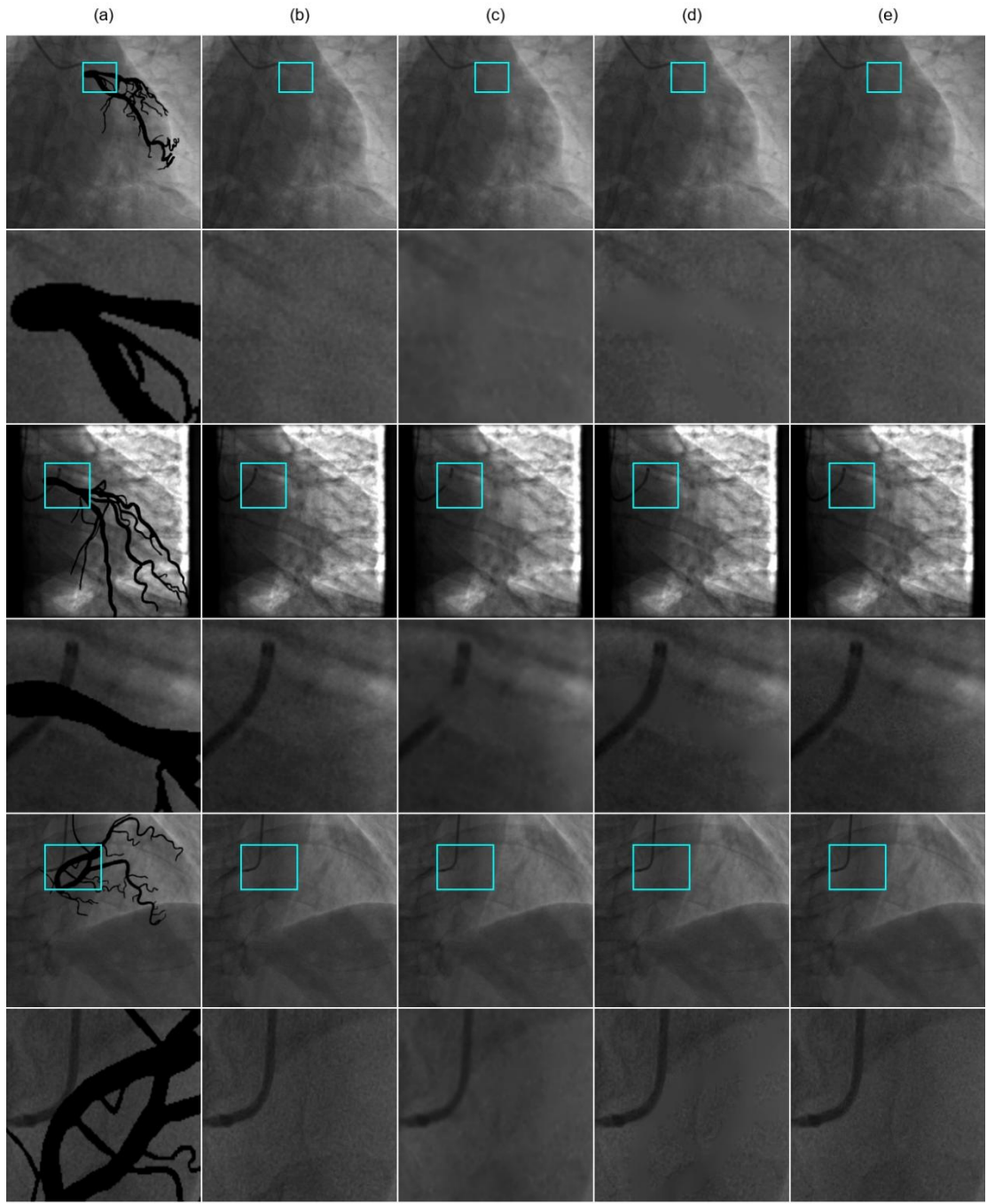

**Figure 5.** Comparison of background restoration results by different networks. (a) synthetic frame; (b) pre-contrast frame; (c) ShadowDiffusion; (d) ShadowFormer; (e) Proposed CBResNet.

## 3. Overall coronary subtraction results

3.1 Qualitative comparison

**Table 2.** Qualitative and quantitative comparison of different background restoration networks.

| Analysis type | Metric | ShadowDiffusion | ShadowFormer | Proposed |
|---|---|---|---|---|
| Qualitative analysis | Overall quality | 4.349 | 4.572 | **4.722** |
| | Restoration fidelity | 4.385 | 4.610 | **4.759** |
| | Transition integrity | 4.498 | 4.684 | **4.796** |
| | Edge artifact presence | 4.386 | 4.572 | **4.684** |
| Quantitative analysis | PSNR | 44.77 | **48.42** | 46.35 |
| | VPSNR | 45.59 | **48.46** | 46.37 |
| | NVPSNR | 52.48 | **74.30** | **74.30** |
| | SSIM | 0.9792 | **0.9918** | 0.9859 |
| | VSSIM | 0.9951 | **0.9975** | 0.9956 |
| | NVSSIM | 0.9988 | **0.9999** | **0.9999** |
| | FID | 41.48 | 8.48 | **4.11** |
| | VFID | 2.18 | 2.52 | **0.48** |

Note: V-prefix denotes vascular regions; NV-prefix denotes Non-Vascular (Background) regions.

Figure 6 presented representative subtraction results. (a) was raw angiogram, (b) was Zeng's DeepSA method, (c) was the proposed CDSA-Net. Our method consistently outperformed the compared approach across all four criteria (Likert scores: vessel clarity 4.72 vs. 4.35, vessel integrity 4.68 vs. 4.30, background suppression 4.73 vs. 4.70, background artifacts 4.66 vs. 4.64). The proposed method achieved superior subtraction results in low-contrast regions, area overlapped with ribs, and regions with complex vascular structures, demonstrating improved preservation of vessel integrity and more accurate extraction of grayscale characteristics.

3.2 Quantitative Evaluation

3.2.1 Segmentation-base metric

Our method achieved higher Dice coefficient (0.894±0.126), significantly outperforming Zeng's method (0.632±0.189). This confirmed that higher-quality subtraction enabled more accurate vessel delineation.

3.2.2 Grayscale fidelity assessment

Since the subtraction operation follows formula 18, the resulting intensities were inversely related to the raw angiographic intensities. Therefore, a negative Pearson correlation was expected, and its absolute value reflected the strength of the linear relationship. The proposed method yielded a higher correlation (-0.640 vs. -0.435), indicating stronger preservation of the relative vascular contrast patterns essential for clinical interpretation. The proposed achieved a substantially lower FID of 48.19 compared to 111.47 for Zeng's method, indicating better perceptual similarity to real angiographic images and demonstrating superior overall image quality.

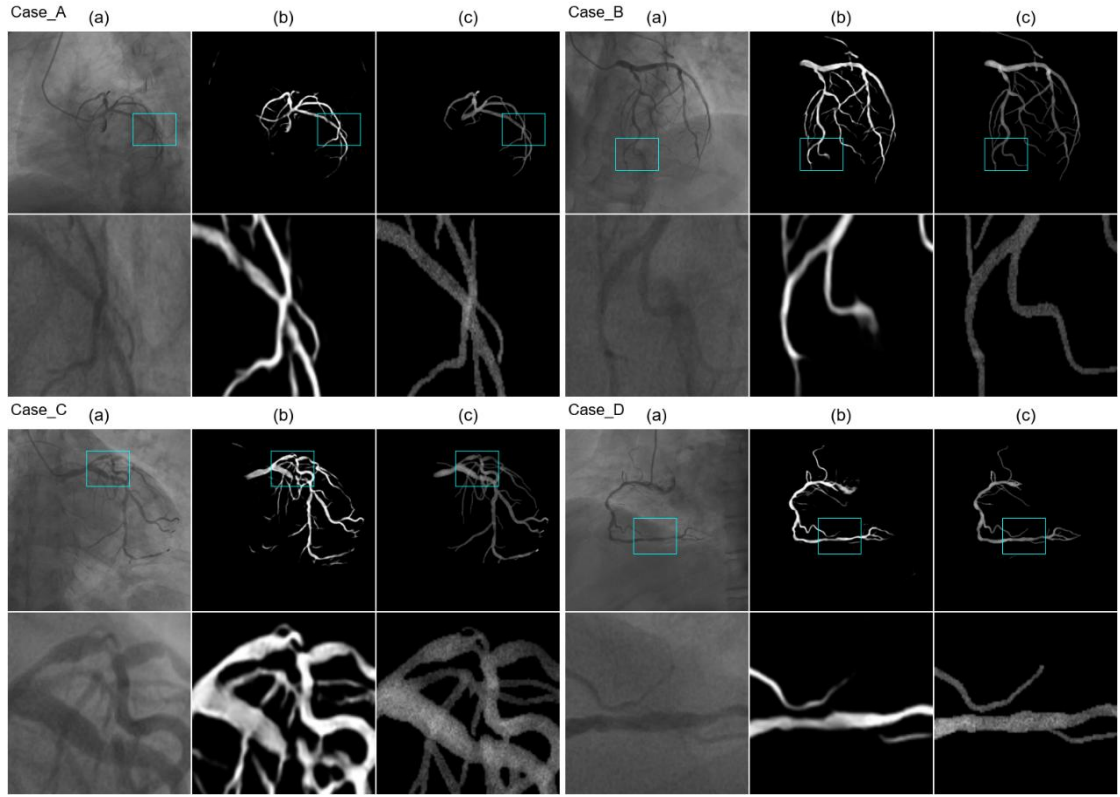

**Figure 6.** Qualitative comparison of coronary subtraction results. (a) raw angiogram; (b) Zeng's DeepSA method; (c) Proposed CDSA method.

## 4. Clinical performance evaluation

Figure 7 and Table 3 summarized the clinical performance of the proposed CDSA method. Figure 7(a) was the stenosis detection by raw and subtracted angiograms. The results visually demonstrated that the subtracted data provided clearer visualization comparing with raw angiograms for localizing stenosis and determining its severity. Figure 7(b) was the TFC assessment by raw and subtracted angiograms. The results demonstrated that the subtracted data enabled more accurate and rapid TIMI frame count (TFC) assessment due to enhanced contrast.

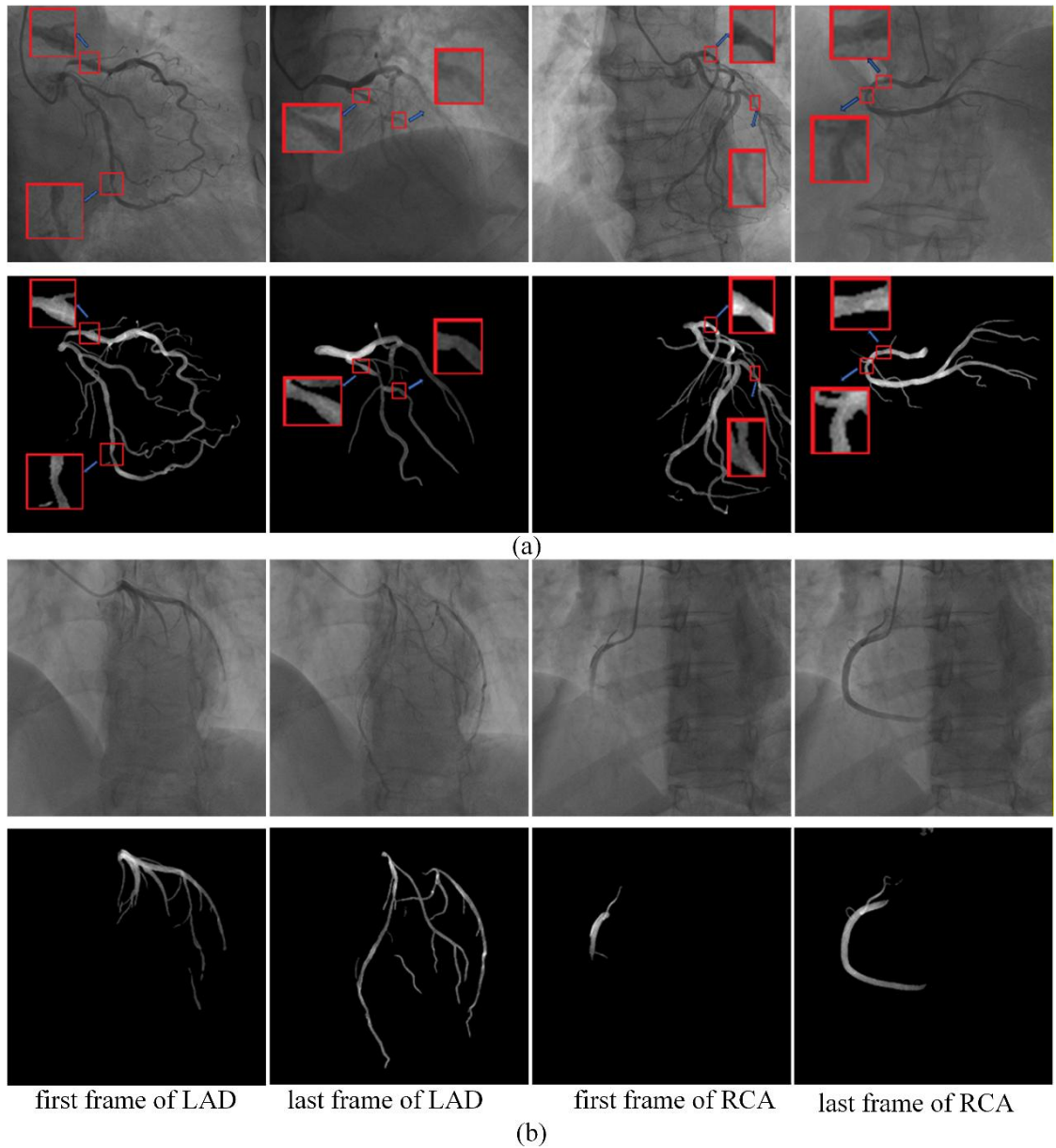

**Figure 7.** Exemplary diagram of clinical application for raw and subtracted angiograms by the proposed CDSA method. (a) stenosis detection by raw and subtracted angiograms; (b) TIMI frame count (TFC) assessment by raw and subtracted angiograms.

Quantitatively, the method showed high agreement with conventional angiography (column 1 to 2 of Table 3). Lesion morphology and hemodynamic analysis indicated clinically acceptable accuracy. Stenosis severity deviations were narrow (left: -25% to 20%, right: -25% to 20%), lesion number discrepancies were minimal (mean differences near zeros), and TFC measurements showed small mean deviations (LAD: -0.503,

Table 3. Clinical performance deviation analysis from the subtracted to the raw angiogram.

| Lesion morphology | | Hemodynamic | Time efficiency(s) |
|---|---|---|---|
| Stenosis severity (deviation range) | Lesion number | TFC value (fps=30) | Stenosis detection /TFC value (percentage relative deviation) |
| (0%, 20%) (left min) (-25%, 0%) (left max) (-20%, 20%) (right min) (-25%, 20%) (right max) | (-0.471±0.920) (left) (-0.331±0.981) (right) | (-0.503±1.332) (LAD) (-0.372±1.201) (RCA) | -25.6%±22.5% / -30.2%±43.8% (TFC LAD) -42.9%±27.7% (TFC RCA) |

Table 4. Clinical subjective evaluation of the proposed CDSA method.

| Qualitative evaluation (5-point Likert scale) | | Diagnostic confidence | Image interpretability | Vessel visibility |
|---|---|---|---|---|
| | Raw data | 3.711±0.332 | 3.653±0.311 | 3.562±0.321 |
| | Segmented data | 3.242±0.253 | 3.542±0.314 | **5.000±0.000** |
| | Subtracted data | **4.010±0.292** | **4.162±0.360** | **5.000±0.000** |

Note: Data in bold represent the optimal values in the comparison.

RCA: -0.372) with standard deviations below 1.34. These results confirmed that subtracted angiograms retained key clinical parameters without systematic bias, supporting their utility in routine workflows.

The method significantly enhanced workflow efficiency (last column of Table 3). Stenosis detection time was reduced by 25.6%, with moderate variability, indicating consistent gains across cases. TFC measurements showed even more pronounced efficiency improvements. These findings confirmed that CDSA-generated angiograms significantly expedited clinical interpretation while maintaining diagnostic reliability.

In a subjective evaluation by cardiologists (Table 4), the subtracted data consistently received the highest ratings for diagnostic confidence (4.010±0.292), image interpretability (4.162±0.360), and vessel visibility (5.000±0.000), outperforming both raw and segmentation-only data. This confirmed that the preserved grayscale information aligned better with clinical reading habits, enhancing diagnostic confidence without disrupting the conventional workflow.

These results collectively validated that the proposed CDSA method maintained diagnostic accuracy while substantially improving visualization, quantitative agreement, and workflow efficiency for clinical coronary analysis.

**Discussions**

**1. Clinical implications**

The proposed CDSA technology streamlines catheterization laboratory workflows and accelerates procedural decision-making during time-sensitive interventions where rapid vessel assessment is paramount.

The subtraction technology removes non-vascular structures (e.g., vertebral shadows, catheter artifacts) while enhancing true vessel signals. It provides artifact-reduced visualization of vessel continuity and stenosis geometry, improving assessment in challenging anatomies and complex flow dynamics, thereby standardizing interpretation across operator experience levels. Additionally, the density-derived information enables quantitative hemodynamic insights beyond anatomical stenosis evaluation.

Clinically, the technology offers several other potential advantages. One is reducing contrast burden. Restoring angiographic clarity in low-visibility scenarios may decrease iodine volume requirements, particularly valuable for renal-impaired patients. Two is enabling superior hemodynamic assessment from subtraction technology. Time-density curves derived from grayscale variations quantify contrast flow velocity, aiding in functional stenosis evaluation without additional pharmacologic stress.

Integration with angiography-derived FFR [4] could establish a unified diagnostic-therapeutic platform, reducing radiation exposure and costs while bridging computational advances with clinical needs to enhance precision in coronary revascularization.

**2. Limitations**

While demonstrating the efficacy in stenosis evaluation, the clinical validation lacked assessment of advanced applications such as contrast kinetics modeling, hemodynam-ic parameter derivation (e.g., angiography-based FFR [4]). Future work could investigate these quantitative functional parameters to fully characterize the technology's impact on procedural optimization and patient outcomes.

## 3. Conclusions

This study established a deep learning-powered coronary subtraction framework that explicitly decouples and synergistically optimizes vessel structure extraction and background restoration, embodying the core philosophy of our CDSA-Net architecture. Validated on clinical angiograms, it delivers high-fidelity vessel extraction while eliminating bony and diaphragmatic obstructions, enabling accurate stenosis assessment even in low-contrast scenarios. The technology expedites procedural workflows, reduces contrast burden, and enhances clinical efficiency, representing a pivotal step toward a computationally augmented catheterization laboratory that standardizes coronary interpretation while prioritizing patient safety.

**Declarations**

**1. Ethic statement**

This study was approved by the Ethics Committee of the Second Affiliated Hospital of Nanchang University. All procedures performed were in accordance with the ethical standards of the institutional research committee and with the 1964 Helsinki declaration and its later amendments. Written informed consent was obtained from all individual participants included in the study. The privacy rights of human subjects have been strictly observed throughout the research process.

**2. Funding**

This work was supported by Special Innovative Projects of Ordinary Colleges and Universities in Guangdong Province [2024KTSCX139]; Research Start-up Project of Guangdong Institute of Science and Technology [2023KYQ182]; Jiangxi Provincial Natural Science Foundation [20232BAB216008]; China Scholarship Council [202506820050].24

## 3. Availability of data and materials

The datasets generated and/or analyzed during the current study are available in the github repository, [https://github.com/DrThink-ai/CDSA-Net].

## 4. Competing interests

The authors declare that they have no competing interests.

## 5. Author Contributions

SL: Data curation, Investigation, Methodology, Software, Validation, Visualization, Writing-original draft. C-KH: Conceptualization, Data curation, Formal analysis, Resources, Writing-original draft. ZHL: Investigation, Software, Visualization. YQH: Conceptualization, Data curation, Formal analysis, Project administration, Resources, Writing-review & editing.